\documentclass[fleqn,usenatbib]{mnras}

\usepackage{newtxtext,newtxmath}

\usepackage[T1]{fontenc}

\DeclareRobustCommand{\VAN}[3]{#2}
\let\VANthebibliography\thebibliography
\def\thebibliography{\DeclareRobustCommand{\VAN}[3]{##3}\VANthebibliography}

\usepackage{graphicx}	
\usepackage{amsmath}	
\usepackage{multirow}
\usepackage{makecell}
\usepackage{adjustbox}
\usepackage{float}

\usepackage{tikz}
\usetikzlibrary{arrows.meta,chains}

\title[Galaxy Morphologies with EfficientNets]{Galaxy Morphology Classification using EfficientNet Architectures}

\author[S. Kalvankar et al.]{
Shreyas Kalvankar,$^{1}$\thanks{E-mail: shreyaskalvankar@gmail.com}
Hrushikesh Pandit,$^{1}$
Pranav Parwate$^{1}$
\\
$^{1}$K. K. Wagh Institute of Engineering Education and Research, Nashik, Maharashtra, India\\
}

\date{Accepted XXX. Received YYY; in original form ZZZ}

\pubyear{2015}

\begin{document}
\label{firstpage}
\pagerange{\pageref{firstpage}--\pageref{lastpage}}
\maketitle

\begin{abstract}
We study the usage of EfficientNets and their applications to Galaxy Morphology Classification. We explore the usage of EfficientNets into predicting the vote fractions of the 79,975 testing images from the Galaxy Zoo 2 challenge on Kaggle. We evaluate this model using the standard competition metric i.e. \textit{rmse} score and rank among the top 3 on the public leaderboard with a public score of 0.07765. We propose a fine-tuned architecture using EfficientNetB5 to classify galaxies into seven classes - completely round smooth, in-between smooth, cigar-shaped smooth, lenticular, barred spiral, unbarred spiral and irregular. The network along with other popular convolutional networks are used to classify 29,941 galaxy images. Different metrics such as accuracy, recall, precision, F1 score are used to evaluate the performance of the model along with a comparative study of other state of the art convolutional models to determine which one performs the best. We obtain an accuracy of 93.7\% on our classification model with an F1 score of 0.8857. EfficientNets can be applied to large scale galaxy classification in future optical space surveys which will provide a large amount of data such as the Large Synoptic Space Telescope. 
\end{abstract}




\section{Introduction}

\subsection{Need for classification of galaxies}

\hspace{0.25 in}Studying galaxies and classifying them into different classes is a long-standing problem. Physicists have been trying to identify and segregate galaxies into individual groups and study their discrete traits to understand the formation of these galaxies and relating the physics that creates them. Morphology is determined by the physical characteristics and the orbital structure of the galaxies. The shape of the galaxy potentials determines which orbital families are present. Stars moving in the specific parts of the space phase generate morphological features such as bars, rings, peanut bulges, pseudo-bulges, etc. Gases pile up close to orbital resonances producing regions of star formation. By looking at the morphology of a particular galaxy, we can learn about the history of star formation and the secular evolution of galaxies. The stability of such features tells us about the distribution of mass throughout the galaxy both dark and luminous.

\subsection{Brief history of classification schemes of Galaxy Morphology}
\subsubsection{\textbf{Classification by visual inspection}}
\hspace{0.25 in}The first classification scheme of galaxy morphology was proposed by Edwin Hubble, 1926. The classification diagram was termed "The Hubble Sequence". The Hubble Sequence (also called "Hubble Tuning Fork") classified the galaxies into three basic classes: spiral, elliptical, irregular \citep{hubble1926extragalactic}.

One of the major challenges in studying morphologies is the limitation of techniques used for measurements that requires an expert human. Before the computerized era of astrophysics, the technique of visual inspection and classification has been used by experts for several decades \citep{hubble1926extragalactic, 1959HDP....53..275D,Edmondson464,1976ApJ...206..883V}. 
Hubble distinguished the galaxies with prominent bulge and disk components. The galaxies with the dominant bulge component were dubbed as Early-Type Galaxies (ETGs) and the galaxies with the prominent disk component were named Late-Type Galaxies (LTGs). ETGs are usually referred to as E-Type or Elliptical Galaxies owing to the lack of structural differentiation and a simpler elliptical shape.
LTGs are commonly known as S-Type or spiral galaxies because of their spiral arms. More refined classification procedures split S-Type galaxies into two groups: Barred and Unbarred spirals. The two groups can be refined further based on their spiral arms' strength. \citet{1963ApJS....8...31D} proposed a number known as T-Type that can be assigned to morphological types. ETGs have T-Type $\le 0$ and LTGs have T-Type$> 0$. Although T-Type considers ellipticity and spiral arms strength, it fails to reflect the bar feature in spirals. The De Vaucouleurs system still retains the basic division of galaxies laid out by Hubble, i.e. Ellipticals, Spirals, Lenticulars and Irregulars. Visually, this system can be represented as a three dimensional version of Hubble's tuning fork.

\citet{Naim/mnras/275.3.567} and \citet{abraham1996b} showed that with limited dynamic range and poor resolution of images, the manual classification leads to large differences among human classifiers. The traditional Hubble sequence begins to breakdown at high redshifts where this discrepancy is mostly produced. The qualitative and environment dependent nature of the hubble classification system limits the quantitative conclusions of how a galaxy's evolution and morphology is related to it's physical parameters. Many physical properties of galaxies have been shown to be correlated and this would expect a quantitative description of how a galaxy's morphology relates to it's physical properties. Many methods of automatic and objective classification have been proposed.

\citet{okamura1994} developed a method that utilized the central concentration index of a galaxy. \citet{abraham1994morphologies} and \citet{abraham1996b} developed a classification system that used the central concentration of light in a galaxy and it's asymmetry. \citet{Conselice_2003} added a 'clumpiness' parameter and related the central concentration, asymmetry and clumpiness parameters to the underlying physical parameters and history. Results from the studies conducted by several authors \citep{1996ApJS..107....1A,Conselice_2000,2004AJ....128..163L} present different objective measures, in addition to those mentioned above, for galaxy morphology namely Smoothness, Gini and M20(CASGM). 

\subsubsection{\textbf{Inception of large scale Galaxy Catalogues}}

\hspace{0.25 in}The computerized era of astrophysics has revolutionized galaxy morphology classification. Both parametrized \citet{1968adga.book.....S, Cohen_2003} and non-parametrized approaches have been used along with combined approaches \citet{2003ApJS..147....1C,2004AJ....128..163L} to reduce each galaxy to one number. This approach enables the processing of large scale images from different sky surveys \citep{Djorgovski_2013} and also helps provide a uniform quantitative set of parameters. 

The previous methods of classification like visual inspection, however effective, were not able to cope with the sheer volume of data provided by the modern sky surveys such as the SDSS (Sloan Digital Sky Survey). This called for a better classification methodology which could process a huge amount of data with much more efficiency. To aid in creation of such systems, some experts have performed extensive work in the detailed visual classification of the subset of the SDSS images. \citet{2007AJ....134..579F} and \citet{refId0} determined modified Hubble types for samples of 2253 and 4458 galaxies respectively. The largest such effort to date, which solely involved researchers, is \citet{2010ApJS..186..427N} which provided detailed classification of 14,034 galaxies. \citet{2008MNRAS.389.1179L} provided classification of nearly one million galaxies by inviting the general public into classifying the galaxies via the internet. This \textit{Galaxy Zoo 1} project obtained more than 40,000,000 classifications made by approx. 100,000 participants. In the next phase \citet{Lintott_2010} included the data release of SDSS consisting of nearly 900,000 galaxies and wrote a paper which presented the measures of classification accuracy and bias. The data from the \citet{2008MNRAS.389.1179L} was substantially reduced by comparing the results with the professional catalogues. Data reduction was performed and no prominent changes were observed after cross-checking. The samples with 80\% agreement among the users were labelled as 'clean' and with 95\% agreement among the users were called 'superclean' samples. The accuracy of the samples, that were created by collecting data clicks and forming them into a scientific catalogue, was very close to the actual morphological feature that the galaxy possessed. For instance, \citet{2010MNRAS.401.1552D}, in their study of merging galaxies, found that
all galaxies with a vote fraction of 40\% or more in the
‘merger’ category were, in fact, true mergers. \textit{The Galaxy Zoo 2} project was launched later and \citet{Willett_2013} produced the data release of nearly 16 million morphological classifications with 304,122 galaxies drawn from the SDSS. While the original Galaxy Zoo project identified galaxies as early-types, late-types or mergers, GZ2 measures finer morphological features \citep{Willett_2013}. This data release allowed a complete study of the finer morphological features and the co-relation of these features with properties of the galaxies i.e. mass, stellar \& gas content, and environment. As mentioned in the paper, although proxies such as spectral features, surface brightness profile, have been used extensively, they cannot be a replacement for full morphological classification as pointed out by \citet{Lintott_2010}. This data has been used in studies of galaxy formation and evolution \citep{Land_et_all_2008,Schawinski,Willett_2015}. 

The third phase in the crowd sourced visual classification was brought forward by \citet{Willet2016HST}. The new \textit{Galaxy Zoo: HST legacy imaging} presented the data release of the Galaxy Zoo: Hubble (GZH). The GZH mainly focused on drawing surveys conducted by the Hubble Space Telescope to view the earlier epochs of galaxy formation. It's data contained measurements of the disc- and bulge-dominated galaxies, spiral disc structure details that relate to Hubble type, identification of bars, and measurements of clump identification. \citet{Willet2016HST} also suggests new methods of calibrating galaxies at different luminosities and at different redshifts by artificially redshfiting the galaxies and using them as a baseline. The present Galaxy Zoo i.e. the 4th incarnation of the project combines new imaging from the SDSS with the most distant images from the Hubble Space Telescope's Cosmic Assembly Near-infrared Deep Extragalactic Legacy Survey (CANDLES) survey. \citet{Simmons2016CANDLES} presented the data release from this survey and provided visual morphologies of approximately 48,000 galaxies observed in
three Hubble Space Telescope legacy fields by the CANDELS. Every galaxy received an average of 40 independent classifications, later combined into detailed morphological information. The information included details on the galaxy features such as clumpiness, bar instabilities, spiral structure, and merger \& tidal signatures. The classification techniques used, preserved classifier independence while effectively down-weighting significantly outlying classifications. \citet{Simmons2016CANDLES} also show that comparing the Galaxy Zoo classifications to any previous classifications of the same galaxies show agreement. This also indicated that the high number of independent classifications provided by Galaxy Zoo, provides an advantage in selecting galaxies with a certain morphology, while combining the data with other classifications, is a more promising approach than using a single, independent method alone.

\subsubsection{\textbf{Classification using Computational Algorithms}}

\hspace{0.25 in}Since the advent of Deep Learning, it has been extensively used in the past two decades in Galaxy Morphology Classification. \citet{Naim/mnras/275.3.567} used Artificial Neural Network classifiers to classify images from the Automated Plate Measurement (APM) Equatorial Catalogue of Galaxies. By using this Supervised approach, the RMS (Root-mean-square) dispersion between the ANN type and correct mean type was comparable to the overall RMS dispersion between the experts.  \citet{Owens/mnras/281.1.153} employed oblique decision trees for the morphological classification using the data from \citet{Storrie1992}. They also indicated that the data could be classified into lesser, but well defined categories. Galaxies could now be confidently classified to larger, overlapping regions, i.e. that multiple decision trees could, in fact, be generated to distinguish easily between different regions along the continuum of classifications. This essentially meant that, while the non-neighbour  classes could be easily separated, the neighbouring classes could not. While trees grown to distinguish \textit{E-type} galaxies from \textit{(Sa+Sb)-types} were accurate, those grown to distinguish \textit{E-types} from \textit{S0-types} would be very inaccurate. \citet{Bazell_2001} used a Naive Bayes classifier and a decision-tree induction algorithm with pruning for automated classification of 800 galaxies, proving that an ensemble of classifiers decreases the classification error. \citet{De_La_Calleja2004} used a neural network along with locally weighted regression method and implemented a homogeneous ensembles of classifiers. It was found that accuracy dropped from 95.11\% to 92.58\% while classifying galaxies into two classes (E and S) and classifying them into three classes (E, S, Irr). Further increase in classes caused the accuracy to drop further i.e. accuracy of 56.33\% for a five-case classification which further reduced to 48.50\% for a seven-case classification \citep{De_La_Calleja2004}. \citet{Banerji_2010} employed Artificial Neural Network to classify  galaxies into three classes i.e. spirals, early types, point sources. A combination of profile fitting and adaptive weighted fitting parameters resulted in better than 90\% accuracy. It was observed that the input parameters are more decisive in achieving greater accuracy than the completeness of magnitude of the training set. \citet{Gauci2010} applied and compared Decision tree algorithms using CART, C4.5, Random Forests and fuzzy logic algorithms. While promising results were achieved in all the employed algorithms, Random Forests gave the highest accuracy.\\
\hspace*{0.25 in}\citet{Ferrari_2015} employed a Linear Discriminant Analysis (LDA) technique to automatically classify galaxies from the astronomical images based on morphometric parameters namely smoothness, asymmetry, concentration, spirality and entropy achieving an accuracy of over 90\%.
\citet{Lecun2015} showed how the performance of classification depends on feature engineering. The feature engineering and feature extraction played an important role in classification models as further models exploited the elemental features of a galaxy image and used advanced algorithms to classify them. Deep learning models consist of multiple non-linear layers that learn data representations and automatically extract features from the raw data that is fed into it \citep{Bengio2013,Lecun2015}. After a series of non-linear transformations, the higher levels have abstract representations of data and these can essentially be used for discrimination and classification purposes. Thus, deep convolutional neural networks (CNNs) have become a particularly dominant approach for image classification and feature extraction. The tremendous datasets, such as the Galaxy Zoo, which are well equipped with the predefined data representations allow faster and efficient implementation of these CNNs, with many works yielding excellent results. \citet{Mairal2014} proposed to train convolutional neural networks to approximate kernel feature maps, which allow the desired invariance properties to be encoded in the choice of kernel, and subsequently be learnt. \citet{Dieleman2015} implemented a 7-layer convolutional architecture for the first time to classify galaxies based on their morphological features by translation of image and exploiting its rotational invariance.

They rotated the images by various angles and turned them into 'viewpoints', then cropped them and fed it to the convolution layers and their output representations were concatenated and processed by a stack of dense layers to obtain the desired predictions. \citet{Huertas} used the CANDELS dataset of over 50,000 galaxies in the 5 CANDELS fields (GOODS-N, GOODS-S, UDS, EGS, and COSMOS) and used the \citep{Dieleman2015} model to classify these high redshift galaxies. \citet{HOYLE} proposed using a base DNN architecture inspired by \citep{Krizhevsky}, that obtains state of the art results on the ImageNet dataset, for estimating the photometric redshift of galaxies. \citet{brunner} proposed a ConvNet inspired from the VGG architecture for classifying stars and galaxies in the SDSS and CFHTLenS photometric images. \citet{dai2018galaxy} proposed a new ConvNet variation inspired from the \citet{Dieleman2015} model and infused it with the ResNets proposed by \citet{he2016} to classify galaxies into 5 classes, achieving promising results.
  
\section{Dataset}
\hspace{0.25 in}After the Galaxy Zoo 1 \citep{2008MNRAS.389.1179L} project was retired, \citet{Willett_2013} came up with the Galaxy Zoo 2 challenge which consisted of $304,122$ galaxy images drawn from the Sloan Digital Sky Survey. The GZ2, along with its predecessor, was a citizen science project which helped with more than 16 million classifications of these images. The original Galaxy Zoo project identified the galaxies into early-types, late-types or mergers. In contrast, the GZ2 measured much finer details and morphological features such as bars, bulges, disks, spirals and more. Essentially, the GZ2 presented quantified data of the said features by providing the strength of bulges or the number of spiral arms. The first model has been developed in the context of the Galaxy Zoo challenge, an online international competition organized by Galaxy Zoo, sponsored by Winton Capital, and hosted on the Kaggle platform for data prediction contests. It was held from December 20th, 2013 to April 4th, 2014. 

The primary sample images consisted of the brightest 25\% of the resolved galaxies from the SDSS North Galactic Cap and images were generated using the SDSS Data Release 7 (DR7) 'Legacy' catalogue \citep{Abazajian2009}. The spectroscopic samples come from the main Galaxy Sample of the SDSS. The goal of the competition was to develop a generalized algorithm that could be applied to many images of the same kind and hence the samples were selected broadly and spread out evenly considering all ranges of colour, size and morphology. The imaging depth of the SDSS and elimination of uncertain and over-represented morphological categories as a function of colour, primarily due to red shifted elliptical galaxies and blue-shifted spiral galaxies, resulted in limited images and hence reduced the data substantially. Elimination of colour as a function helped to ensure that the colour of galaxies was not used as a proxy for morphologies and any high performance model would solely use the structural parameters of a galaxy. 

The final training set consisted of 61,578 JPEG images in RGB and the probabilities for each of the 37 answers in the GZ2 decision tree. The evaluation set consisted of 79,975 images of the same kind but no morphological data was provided. Each image is 424 by 424 pixels comprising of 3 channels. The probabilities are the actual processed vote fractions that have been obtained from the answers of the GZ2 participants. We treat these vote fractions as probabilities in the paper. The goal was to predict these probabilities values for every image in the evaluation set with the highest precision. The morphological data provided in the training set is a modified version of the vote fractions in the GZ2 catalogue. These vote fractions were transformed into cumulative probabilities that assigned higher weights to the more fundamental morphological categories higher in the decision tree. Images were anonymized based on their SDSS IDs. The goal, being predicting probabilities rather than determining the most likely answer for each question in the decision tree, makes this a \textit{regression} problem and not a \textit{classification} problem. The performance of the model was evaluated by computing root-mean-squared error (RMSE) between the predicted values and the corresponding probabilities derived from the vote fraction values. Let $p_n$ be the answer probability values for an image and $p'_n$ be the predicted values (n = 1...37). The RMSE($p,p'$) can be given by

\begin{equation}
    rmse(p,p') = \sum_{n=1}^{37}\sqrt{p^2-p'^2}
    \label{RMSE_eq}
\end{equation}

The metric was chosen because it puts more emphasis on questions with higher answer probabilities. 
\begin{footnotesize}
\begin{table}
\resizebox{\linewidth}{!}{
\begin{tabular}{c p{5 cm} l r} 
\hline
 Task & Question & Responses & Next\\
\hline
\hline
\multirow{3}{*}{01} & \multirow{3}{*} {\thead[l]{Is the galaxy simply smooth\\ and rounded, with no sign of\\a disk?}} & smooth & 07\\
& & features or disk & 02\\
&  & star or artifact & \textbf{end}\\
\hline
\multirow{2}{*}{02} & \multirow{2}{*} {\thead[l]{Could this be a disk viewed\\ edge-on?}} & yes & 09\\
& & no & 03\\
\hline
\multirow{3}{*}{03} & \multirow{3}{*} {\thead[l]{Is there a sign of a bar\\feature through the centre\\of the galaxy?}} & yes & 04\\
& & no & 04\\
\vspace{1\baselineskip}\\
\hline
\multirow{2}{*}{04} & \multirow{2}{*} {\thead[l]{Is there any sign of a\\spiral arm pattern?}} & yes & 10\\
& & no & 05\\
\hline
\multirow{4}{*}{05} & \multirow{3}{*} {\thead[l]{How prominent is the\\central bulge, compared\\with the rest of the galaxy?}} & no bulge & 06\\
& & just noticible & 06\\
& & obvious & 06\\
& & dominant & 06\\
\hline
\multirow{2}{*}{}06 & \multirow{1}{*} {\thead[l]{Is there anything odd?}} & yes & 08\\
& & no & \textbf{end}\\
\hline
\multirow{3}{*}{}07 & \multirow{1}{*} {\thead[l]{How rounded is it?}} & completely round & 06\\
& & in between & 06\\
&  & cigar-shaped & 06\\
\hline
\multirow{7}{*}{}08 & \multirow{3}{*} {\thead[l]{Is the odd feature a ring,\\or is the galaxy distributed\\or irregular?}} & ring & \textbf{end}\\
& & lens or arc & \textbf{end}\\
& & distributed & \textbf{end}\\
& & irregular & \textbf{end}\\
& & other & \textbf{end}\\
& & merger & \textbf{end}\\
& & dust lane & \textbf{end}\\
\hline
\multirow{3}{*}{09} & \multirow{3}{*} {\thead[l]{Does the galaxy have a\\bulge at its centre? If\\so, what shape?}} & rounded & 06\\
& & boxy & 06\\
&  & no bulge & 06\\
\hline
\multirow{3}{*}{}10 & \multirow{2}{*} {\thead[l]{How tightly wound do the\\spiral arms appear?}} & tight & 11\\
& & medium & 11\\
&  & loose & 11\\
\hline
\multirow{6}{*}{}11 & \multirow{2}{*} {\thead[l]{How many spiral arms\\are there?}} & 1 & 05\\
& & 2 & 05\\
& & 3 & 05\\
& & 4 & 05\\
& & more than four & 05\\
& & can't tell & 05\\
\hline
\end{tabular}}
    \caption{The GZ2 decision tree, comprising of 11 tasks and 37 responses. Reproduced from Table 2 in \citet{Willett_2013}}
    \label{GZ2_DecisionTree}
\end{table} 
\end{footnotesize}
\begin{center}
    \begin{table*}
\centering
    \begin{tabular}{c l c l c}
    \hline
    \hline
        Class & Sample & Tasks & Selection & $N_{sample}$\\
    \hline
        \multirow{3}{*}{}0& Completely round smooth & T01 & $f_{smooth}\geq0.469$ & 8107\\
        & & T07 & $f_{completely round}\geq0.5$ & \\
        & & T06 & $f_{odd/no}\geq0.5$ & \\
    \hline
        \multirow{3}{*}{}1& In-between smooth & T01 & $f_{smooth}\geq0.469$ & 7782\\
        & & T07 & $f_{in-between}\geq0.5$ & \\
        & & T06 & $f_{odd/no}\geq0.5$ & \\
    \hline
        \multirow{3}{*}{}2& Cigar shaped smooth & T01 & $f_{smooth}\geq0.469$ & 578\\
        & & T07 & $f_{cigar-shaped}\geq0.5$ & \\
        & & T06 & $f_{odd/no}\geq0.5$ & \\
    \hline
        \multirow{3}{*}{}3& Lenticulars & T01 & $f_{features/disk}\geq0.430$ & 3780\\
        & & T02 & $f_{edge-on/yes}\geq0.602$ & \\
        & & T06 & $f_{odd/no}\geq0.5$ & \\
    \hline
        \multirow{4}{*}{}4& Barred spirals & T01 & $f_{feature/disk}\geq0.430$ & 827\\
        & & T02 & $f_{edge-on/no}\geq0.715$ & \\
        & & T03 & $f_{bar/yes}\geq0.715$ & \\
        & & T04 & $f_{spiral/yes}\geq0.619$ & \\
    \hline
        \multirow{4}{*}{}5& Unbarred spirals & T01 & $f_{feature/disk}\geq0.430$ & 3307\\
        & & T02 & $f_{edge-on/no}\geq0.715$ & \\
        & & T03 & $f_{bar/no}\geq0.715$ & \\
        & & T04 & $f_{spiral/yes}\geq0.619$ & \\
    \hline
        \multirow{6}{*}{}6& Irregular & T06 & $f_{odd/yes}\geq0.420$ & 1560\\
        & & T08 & $f_{disturbed|irregular|other|merger|dustlane}\geq0.5$ & \\
    \hline
    \end{tabular}
    \label{Classes}
    \caption{Clean samples selected from GZ2 data. The thresholds from \citet{Willett_2013} are used for well-sampled images. The thresholds depend upon the number of votes for a classification task considered to be sufficient.}
\end{table*}
\end{center}
\hspace{0.25 in}The first part of this study uses the data provided by the GZ2 project competition. The later part of the study actually obtains clean samples of the galaxies belonging to specific morphologies depending on their appropriate thresholds. These thresholds depend on the number of votes for a classification task considered to be sufficient \citep{dai2018galaxy}. These thresholds are considered conservative for the selection of clean samples in \citep{Willett_2013}. \citet{dai2018galaxy} used the methodology to classify galaxies into five classes namely, completely round smooth, in-between smooth(between completely round and cigar-shaped), cigar-shaped smooth, edge-on and spiral. Following the same, we classify the galaxies into seven classes i.e. completely round smooth, in-between smooth, cigar-shaped smooth, lenticular (edge-on), barred spiral, unbarred spiral and irregular (irregular set is a collection of mergers, disturbed, dust lanes and irregular galaxies). The thresholds for smooth galaxies have been loosened from 0.8 to 0.5 \citep{dai2018galaxy}. For complete details, refer \citet{Willett_2013}. Table~\ref{Classes} contains the clean sample selection criterion for every class. The classes have been labelled from 0,1,...,6. Every class contains 8107, 7782, 578, 3780, 827, 3307 and 1560 samples respectively. The dataset reduces to 25,941 images that are divided in a train, test ratio of 9:1. Thus, we end up with 23,352 training images and 2589 testing images.
\begin{center}
    \begin{table}
\scriptsize
    \begin{tabular}{c c c c c c c c c }
    \hline
    \hline
        & 0 & 1 & 2 & 3 & 4 & 5 & 6 & Total\\
    \hline
    \hline
        \multirow{3}{*}{}Training Set & 7297 & 7004 & 521 & 3402 & 745 & 2979 & 1404 & 23352 \\
        
         \multirow{3}{*}{}Testing Set & 810 & 778 & 57 & 378 & 82 & 328 & 156 & 2589 \\

        \hline
        \multirow{3}{*}{}Data Set & 8107 & 7782 & 578 & 3780 & 827 & 3307 & 1560 & 25941 \\
    \hline
    \hline
    \end{tabular}
    \label{Dataset}
    \caption{Number of galaxy images in each morphological class 0, 1, 2, 3, 4, 5, 6 (as represented in the Table 2)}
\end{table}
\end{center}
\begin{figure}
    \centering
    \includegraphics[width=0.4\textwidth]{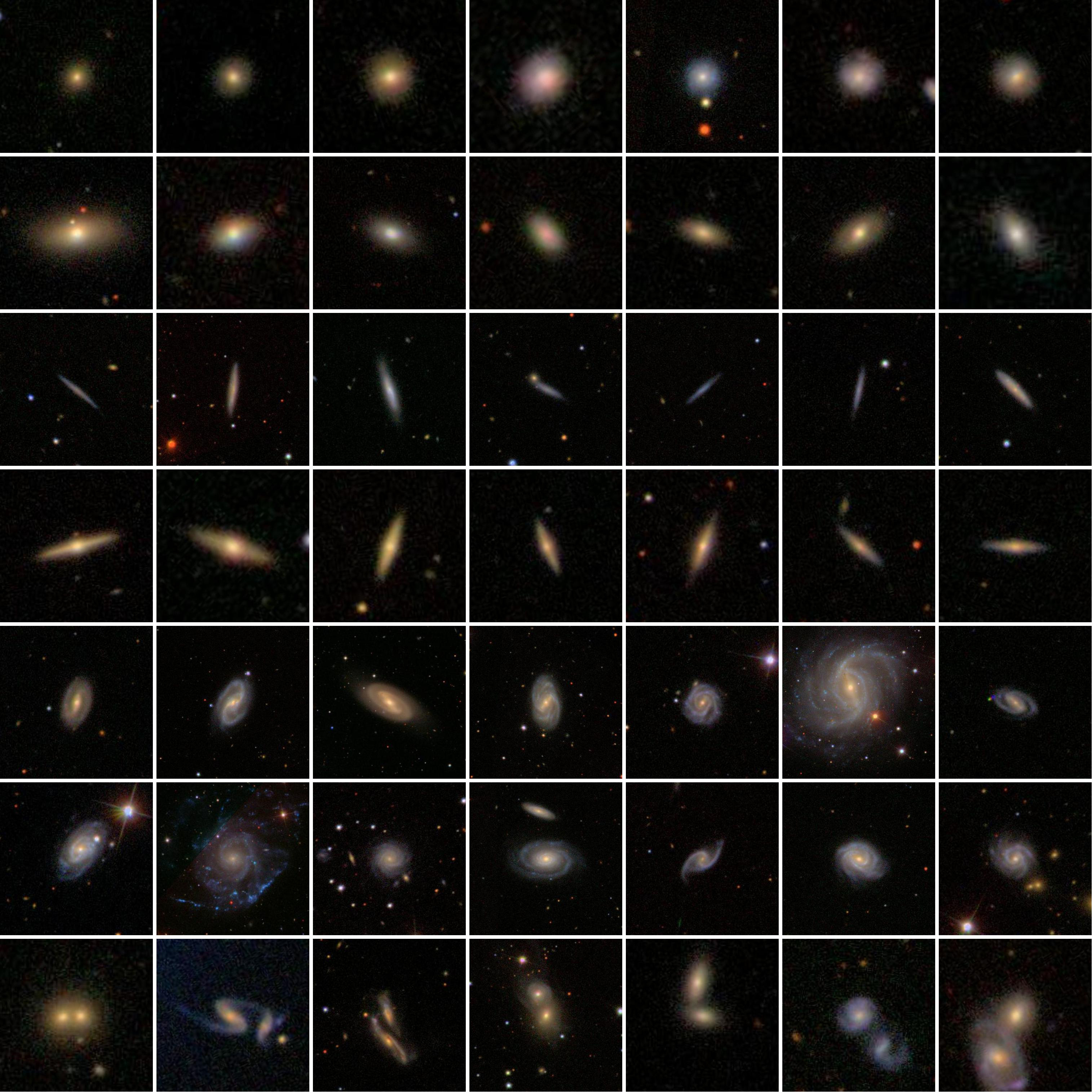}
    \caption{Example galaxy images from the dataset. Each row represents a class. From top to bottom, their Galaxy Zoo 2 labels are class 0, 1, 2, 3, 4, 5 and 6}
    \label{sample_images}
\end{figure}
\hspace{0.25 in}Figure \ref{sample_images} shows samples of galaxies belonging to classes 0,1,2,3,4,5,6 respectively. Table \ref{Dataset} shows the number of samples in train set and test set for each class. 

\section{Neural Networks}
\subsection{Artificial Neural Networks}
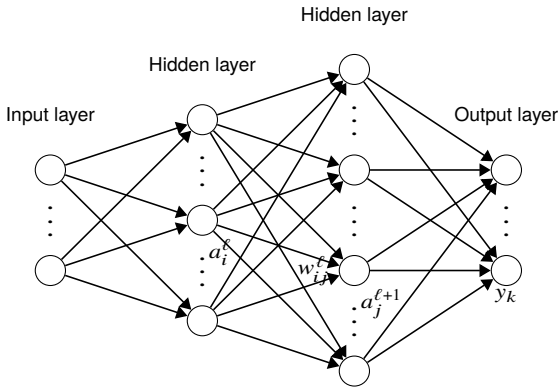
\begin{figure}
    \centering
    \begin{tikzpicture}[neuron/.style={circle,inner sep=.5em,draw},
    neuron missing/.style={
    scale=1.25,
    text height=0.333cm,
    execute at begin node=$\vdots$,
  }]
 \begin{scope}[x=2cm,y=1.333cm]  
 \foreach \X/\Y [count=\Z,remember=\Y as \LastY] in {Input/2,Hidden/3,Hidden/4,Output/2} 
 {\foreach \YY in {1,...,\Y}
 {\node[neuron] (n-\Z-\YY) at (\Z,-\YY+\Y/2+1/2) {};
 \ifnum\YY=1
  \node[above=1em of n-\Z-\YY,font=\sffamily] {\X\ layer};
 \else
  \path (n-\Z-\the\numexpr\YY-1) -- (n-\Z-\YY) 
   node[midway,neuron missing] {};
 \fi
 \ifnum\Z>1
  \foreach \YYY in {1,...,\LastY}
  {\draw[-{Triangle},semithick] (n-\the\numexpr\Z-1\relax-\YYY) -- (n-\Z-\YY) ;}
 \fi
 }}
 \end{scope} 
 \path[nodes={inner sep=1pt}]
    (n-2-2.-80) node[below right]{$a_i^\ell$} 
    (n-3-3.-80) node[below right]{$a_j^{\ell+1}$}
    (n-3-3.west) node[left=2pt]{$w_{ij}^{\ell}$}
    (n-4-2.-90) node[below]{${y_k}$};
\end{tikzpicture}
    \caption{A feed forward neural network}
    \label{fig: ANN}
\end{figure}

\hspace{0.25 in}The concept of Artificial Neural Networks was brought forth by \citet{mcculloch1943logical} where they described the behaviour of a neuron using rules of calculus and created a computational model for neural networks. Neural networks are computing systems that are inspired from the biological neural networks that exist in animal brain. Rochester was the first to simulate a neural network in the IBM Research Laboratory. These networks are made of adaptive units that are interconnected to stimulate a response to real world objects. Artificial Neural Networks perform tasks by considering examples rather than being programmed with task-specific rules. Figure \ref{fig: ANN} shows a schematic of the simplest multi-layer perceptron network, i.e. a feed-forward neural network. The network is composed of an input layer, hidden layers and output layer. Define $a_i^\ell$ as the $i^{th}$ neuron of the $\ell^{th}$ layer and $a_j^{\ell+1}$ as the $j^{th}$ neuron of the $(\ell+1)^{th}$ layer and $w_{ij}^\ell, b_j^\ell$ is the weight and bias connecting the two neurons, then the output of the $\ell^{th}$ layer is given by:
\begin{equation}
    a^{\ell+1}_j = g(\sum_{i \in N^\ell}(w_{ij}^{\ell}a_i^{\ell} + b_j^{\ell}))
    \label{ANN_activation}
\end{equation}

The loss is calculated using a loss function such as cross entropy function especially for binary classification.
\begin{equation}
    loss(y',y) = -y\log{y'} - (1-y)\log{(1-y')}.
    \label{ANN_loss}
\end{equation}
where $y' \in \{0,1\}, y \in (0,1)$. This objective is to minimize this cross entropy over a batch of all training data. This is done using gradient descent where the parameters viz. weights and biases are updated to reduce the overall cross entropy loss.

\subsection{Convolutional Neural Networks}
\hspace{0.25 in}Convolutional Neural Networks \citep{LeCun1989} is a class of deep neural networks that are applied to images. They are translation invariant and are called shift invariant or space invariant artificial neural networks. The Convnets (also called CNNs) are created to process multiarray data. The network takes an input image and assigns importance to the various aspects of the object in the image and then differentiates one object from other. As Convnets are able to successfully capture the spatial and temporal dependancies in an image by using different filters, they perform better on an image data set. Due to this, CNNs have become reliable and successful in practical applications. A CNN has the same architecture as that of an ANN, an input layer, an output layer and hidden layers. The hidden layer of a CNN performs a series of operations that convolve with a dot product. A classic CNN layer consists of three stages \citep{Goodfellow-et-al-2016}. First, the layer performs several convolutions. Then, a non-linear activation function is applied and finally, a pooling method modifies the output. Convnets contain convolutional layers, pooling layers and fully connected blocks.\\
\hspace*{0.25 in}\textbf{Convolutional:} A convolution is a linear operation that can be viewed as a multiplication or dot product of matrices. The input is a tensor of shape $height$ x $width$ x $channels$ and the convolution operation abstracts the image to a feature map (also called a kernel) of shape $kernel size$ x $kernel size$ x $kernel channels$. The layers can be computed by:
\begin{equation}
    a^{\ell+1}_j = g(\sum_{i \in F_j}(a_i^\ell \times k_{ij}^{\ell+1} + b_j^{\ell+1}))
    \label{Convolution}
\end{equation}
where $\ell$ is the layer, g is the activation function ReLU, $F_j$ is the receptive field, k is the convolutional kernel and b is the bias. \\
\hspace*{0.25 in}\textbf{Pooling:} The pooling layer performs subsampling on the data. It includes local or global pooling. The pooling layer reduces the dimensions of the output by combining the neuron clusters into one neuron. It can be done by computing max or average of the clustered neurons.\\
\hspace*{0.25 in}\textbf{Receptive field:} In CNN, a neuron receives input from a restricted sub area of the previous layer as opposed to a fully connected block. This subarea is generally of a square shape. The input area of a neuron is called as its receptive field.\\
\hspace*{0.25 in}\textbf{Fully Connected layer:} The output of the last convolutional layer or pooling layer is typically connected to a fully connected block. Fully connected block has layers that connect each neuron to every neuron in the previous layer and is essentially the same as the multi-layer perceptrons proposed by \citet{mcculloch1943logical}.\\
\hspace*{0.25 in}While shallow neural networks do well in memorization, more deeper architectures are needed to ensure generalization,they learn high-level features from the input data in an incremental manner. Multiple Layers can learn features at various levels of abstraction. Deeper Neural Networks tend to have the problem of overfitting due to having a vast number of parameters. To deal with this problem, new methodologies like ReLu \citep{Nair}, Dropout \citep{srivastava}, Data augmentation \citep{Krizhevsky}, Batch Normalization \citep{Ioffe}. Pioneering work in deepeer neural networks was done with the introduction of Alexnet by \citep{alex}. It was followed by ZFNet \citep{zeiler} which was a similar architecture to Alexnet. Inception or GoogLeNet \citep{Szegedy}, VGG \citep{simonyan}, Resnet \citep{he2016} and DenseNet \citep{Huang} were all followed shortly. The latest addition to this was done by \citep{tan2019efficientnet} who introduced the Efficient Nets.

\subsection{Residual Networks}
\hspace*{0.25 in}The increasing popularity of Convnets called for a need of more efficient architectures for a wide sprectrum tasks. This led on to various ventures that introduced us with a variety of different networks, for e.g. the Visual Geometric Group's weighted layer networks \citep{simonyan}. The Large Scale Visual Recognition Challenge classification contest (LSVRC2012) paved way for the a new dense CNN that was the AlexNet \citep{alex}. The celebrated victory of Alexnet spearheaded a revolutionary approach of designing deeper CNNs with a foundation for the traditional convolution block of a convolutional layer followed by an activation function and then a max pooling operation. The success of the CNNs has been accredited to these additional layers with the intuition being that the network progressively learns more complex shapes with increasing layers. However, \citet{he2016} empirically showed that there is a limit to the number of layers a network can have. He proved that there exists a threshold for depth in a traditional CNN model. This particular drawback inhibited and constrained the development of CNNs until \citet{he2016} came up with another groundbreaking architecture that will last a few years in the computer vision community. With his Residual blocks, the possibility of building a network that could have the depth of a thousand layers was made possible. Residual neural networks (also called ResNets) is a class of ANNs that closely resemble the constructs of pyramidal cells in the cerebral cortex of the brain. The failure of deeper traditional CNNs could be attributed to numerous factors such as the initialization of network, optimization functions or the vanishing gradients. \citet{he2016} proposed the framework where the layers try to learn from a residual mapping instead of the underlying mappings of few layers. \\
\begin{figure}
    \centering
    \begin{tikzpicture}[box/.style={draw,thick,minimum width=5em,minimum
    height=1.2em},arj/.style={semithick,-latex}]
  \begin{scope}[start chain=A going below,oc/.style={join=by arj,on chain},
    font=\sffamily,node distance=1.75em]
   \node[box,oc]{Layer 1};
   \node[box,oc,label={north west:{$F(a_\ell)$}}]{Layer 2};
   \node[circle,oc,draw,thick,node font=\bfseries\large,inner sep=1pt,
    label={left:{$F(a_\ell)+a_\ell$}}]{+};
   \node[oc]{$a_{\ell+1}$};
  \end{scope}   
  \draw[arj,latex-] (A-1.north) -- ++ (0,1.75em)
      coordinate[midway,label=left:{$a_\ell$}](aux);
  \draw[arj] (aux) -- ++ (2em,0) to[out=0,in=0,looseness=1.5]
  node[midway,right]{$h(a_\ell)=a_\ell$}(A-3);    
\end{tikzpicture}
    \caption{Residual Block}
    \label{fig: residual block}
\end{figure}
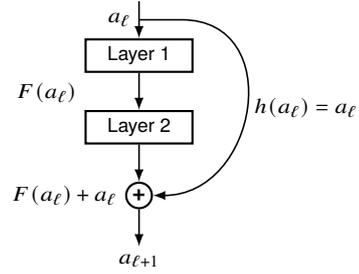
\hspace*{0.25 in}Figure \ref{fig: residual block} shows a building block of a residual network. The residual blocks can be expressed mathematically as follows. Let $h(a)$ be an underlying mapping that is to be fit by a set of layers, where $a$ is the input to these layers. If we hypothesize that multiple non-linear transformations by these layers can approximate the layer functions, then one can also hypothesize that a similar approximation can be made for the residual functions, i.e. $h(a) - a$, which have the same input and output dimensions. So instead of letting the underlying mapping be $h(a)$ we approximate a residual function $F(a) = h(a) - a$. Thus, the actual function becomes $h(a) = F(a) + a$. We can achieve this approximation in a feed forward neural network using a series of skip connections that perform identity mapping and jump over a few layers. Adding the outputs of these two connections gives the final output layer. Thus, the residual unit can be defined as,
\begin{equation}
    a_{\ell+1} = h(a_\ell) + F(a_\ell, W_\ell)
    \label{Residual unit}
\end{equation}
where $a_{\ell+1}$ and $a_\ell$ represent the input and output for the $\ell^{th}$ layer and $F$ is the residual function. The $h$ denotes the operation in the identity mapping. The output of the layer is generally processed using an activation function such as ReLU. Let $f$ be the ReLU activation and replacing $F$ with the definition of feed forward activation, the residual block thus can be defined as, 
\begin{equation}
    a_{\ell+1} = f(h(a_\ell) + W_2f(W_1a_\ell))
    \label{Residual unit redefined}
\end{equation}
For the sake of simplicity, we consider no operation being performed in the identity mapping. Thus, equation \ref{Residual unit} and equation \ref{Residual unit redefined} become,
\begin{equation}
    a_{\ell+1} = f(a_\ell + F(a_\ell, W_\ell))
    \label{Final Residual notation}
\end{equation}
\subsection{EfficientNets}
\hspace{0.25 in}With the introduction of ResNets \citep{he2016}, scaling up became prominent in ConvNets to achieve better accuracy. \citet{he2016} demonstrated that by using a residual mapping, it was possible to go deeper and deeper with convolutional architectures. For e.g. ResNet18 was scaled up to ResNet200 by simply adding more layers. However popular, the process of scaling up is poorly understood. While \citet{he2016} proposed scaling up using depth using the famous residual blocks to solve the exploding gradients problem, \citet{Zagoruyko_2016} suggested another way of scaling up using width. With increasing depth, the network accuracy keeps increasing, but every fraction of an increase in accuracy results from doubling the number of layers. Thus, these networks have a problem of diminishing feature reuse and that makes them very slow to train \citep{Zagoruyko_2016}. Taking this into consideration, \citet{Zagoruyko_2016} took the ResNet architecture and increased the width while reducing the depth of the network. With these kinds of scaling, it was evident that one factor was undermined while the other was in focus. \citet{tan2019efficientnet} came up with an approach that balanced all the dimensions of the network i.e. width, height, depth, resolution by scaling them with a constant ratio. This method of compound scaling uniformly scaled every feature with a fixed set of scaling coefficients which gave rise to the \textit{EfficientNets}. 
\subsection{Squeeze and Excitation Networks}
\hspace{0.25 in}Convolutional Networks consist of convolutional layers which has a collection of filters that express the neighbouring spatial connectivity patterns along the input channel. These inturn fuse the channel-wise information together within local receptive fields \citep{hu2018squeeze}. A widely sought after effect for these networks is that these convolutions followed by a series of non-linear activations will produce strong image representations to capture only the defining properties of the image. \citet{hu2018squeeze} proposed a new architectural unit that explicitly modelled the dependencies between channels of its convolutional filters. This approach enabled feature re-calibration through which the network can learn to use global information to suppress less useful features and selectively emphasize on informative features of an image. This block was termed the \textit{Squeeze-And-Excitation Block} or \textit{SE}-block.

For a transformation $F_{tr}$, mapping over an input $X \in {\rm I\!R}^{H'\times W' \times D'}$ to a feature maps $M\in {\rm I\!R}^{H\times W \times D}$. Here, $F_{tr}$ is the convolutional operator and we denote learnt set of filter kernels as $V = [v_1, v_2, ..., v_D]$ where $v_i$ refers to the parameter of the $i$-th filter. The output $U = [u_1, u_2, ..., u_D]$ can be represented by,
\begin{equation}
    u_i = v_i \ast X = \sum_{j=1}^{D'}{v_i^j \ast x^j}
    \label{SE convolutional eqn}
\end{equation}
\hspace{0.25 in}The $\ast$ denotes convolution. $v_i$, $X$ and $u_i$ $\in {\rm I\!R}^{H\times W}$. $v_i^j$ is a 2D spatial kernel acting on the corresponding channel of X. 
\subsubsection{Squeeze: Global Information Embedding}
\hspace{0.25 in}To exploit the channel dependencies, \citet{hu2018squeeze} came up with the \textit{Squeeze} block where signal to each channel in the output feature was considered. As each filter operates using a local receptive field, each unit of transformation output U cannot learn anything outside this region. The \textit{Squeeze} block was proposed to diminish this problem. This block essentially squeezes global spatial information into a single channel. This is done using global average pooling to obtain channel-wise statistics. A descriptor $g \in {\rm I\!R}^D$ is generated by shrinking the output U through its spatial dimensions such that $i$-th element is obtained by:
\begin{equation}
    \centering
    g_i = F_{sq}(u_i) = \frac{1}{H \times W} {\sum_{j=1}^{H}\sum_{k=1}^{W}u_i(j,k)}
\end{equation}
\begin{figure}
    \centering
    \begin{tikzpicture}[box/.style={draw,thick,minimum width=5em,minimum
    height=1.2em},arj/.style={semithick,-latex}]
  \begin{scope}[start chain=A going below,oc/.style={join=by arj,on chain},
    font=\sffamily,node distance=1.75em]
   \node[box,oc]{Squeeze};
   \node[box,oc,label={north west:{$z=F_{sq}(U)$}}]{Reshape, 1 x 1 x D};
   \node[box,oc,label={north west:{$F_{ex}(z)$}}]{Conv, 1 x 1 x m};
   \node[box,oc,label={north west:{$F_{ex}(z)$}}]{Conv, 1 x 1 x D};
   \node[box,oc,label={north west:{$s=F_{ex}(z)$}}]{ReLU};   
   \node[circle,oc,draw,thick,node font=\bfseries\large,inner sep=1pt,
    label={left:{$F_{scale}(U,s)$}}]{x};
   \node[oc]{$\tilde{X}$};
  \end{scope}   
  \draw[arj,latex-] (A-1.north) -- ++ (0,1.75em)
      coordinate[midway,label=left:{$U_{H \times W \times D}$}](aux);
  \draw[arj] (aux) -- ++ (2em,0) to[out=0,in=0,looseness=1.5]
  node[midway,right]{$U$}(A-6);    
\end{tikzpicture}
    \caption{An SE block using pointwise convolutions implemented in EfficientNets}
    \label{SE Block}
\end{figure}
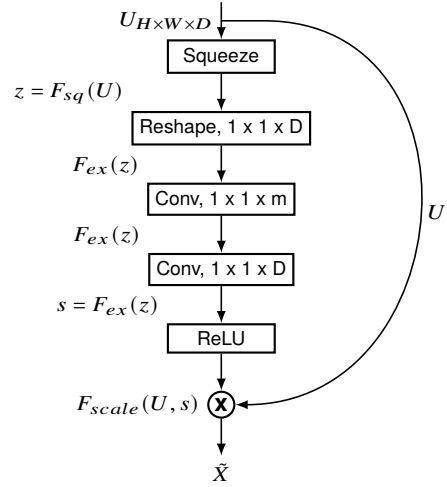
\subsubsection{Excitation: Adaptive Recalibration}
\hspace{0.25 in}After embedding the global information by \textit{squeezing} operation, the second task was to fully capture channel-wise dependencies. To do this, the network is supposed to be flexible i.e. it is capable of learning through non-linear interactions between layers and it should learn to emphasize multiple channel by learning a non-mutually exclusive relationship rather than a one-hot activation. To do this, \citet{hu2018squeeze} opted for a simple FC block by employing a gating mechanism with a sigmoid activation:
\begin{equation}
    s = F_{ex}(z, W) = \sigma(g(z, W)) = \sigma(W_2 \delta(W_1z))
\end{equation}
where $\delta$ refers to the ReLU activation, $W_1 \in {\rm I\!R}^{\frac{D}{r} \times D}$ and $W_2 \in {\rm I\!R}^{D \times \frac{D}{r}}$ The bottleneck from $D\Rightarrow \frac{D}{r}$ limits the model complexity and aids generalization.

To further reduce the model complexity, we opt for the EfficientNet variant of the SE block which uses point-wise convolutions to reduce the trainable parameters as shown in fig.\ref{SE Block}. The new gating mechanism is given by:
\begin{equation}
    s = F_{ex}(z, W) = \delta(\xi(z, W)) = \delta(W_2 \xi (W_1, z))
\end{equation}
where $\delta$ refers to the ReLU activation, $\xi$ corresponds to pointwise convolutional operation, $W_1 \in {\rm I\!R}^{(D+1) \times m}$ and $W_2 \in {\rm I\!R}^{(m+1) \times D}$. As we can observe, the bottleneck from $D \Rightarrow m$ and expansion from $m \Rightarrow D$ substantially reduces the number of parameters from a factor of $D^2$ to $D$.

The final output block is obtained by rescaling $U$ with the activations $s$:
\begin{equation}
    \Tilde{x_i} = F_{scale}(u_i,s_i) = s_i \times u_i
\end{equation}
where $\Tilde{X}=[\Tilde{x_1}, \Tilde{x_2}, ...,\Tilde{x_D}]$ and $F_{scale}(u_i,s_i)$ is channel-wise multiplication of scalar $s_i$ and feature map $u_i \in {\rm I\!R}^{H \times W}$.

\section{Approach}
\hspace{0.25 in}In the previous section, we describe the theory of ResNets, EfficientNets and the Squeeze \& Excitation block. In this section, we present our framework for both the networks with data preprocessing, data augmentation and network architecture and implementation details.
\begin{figure*}
    \centering
    \input{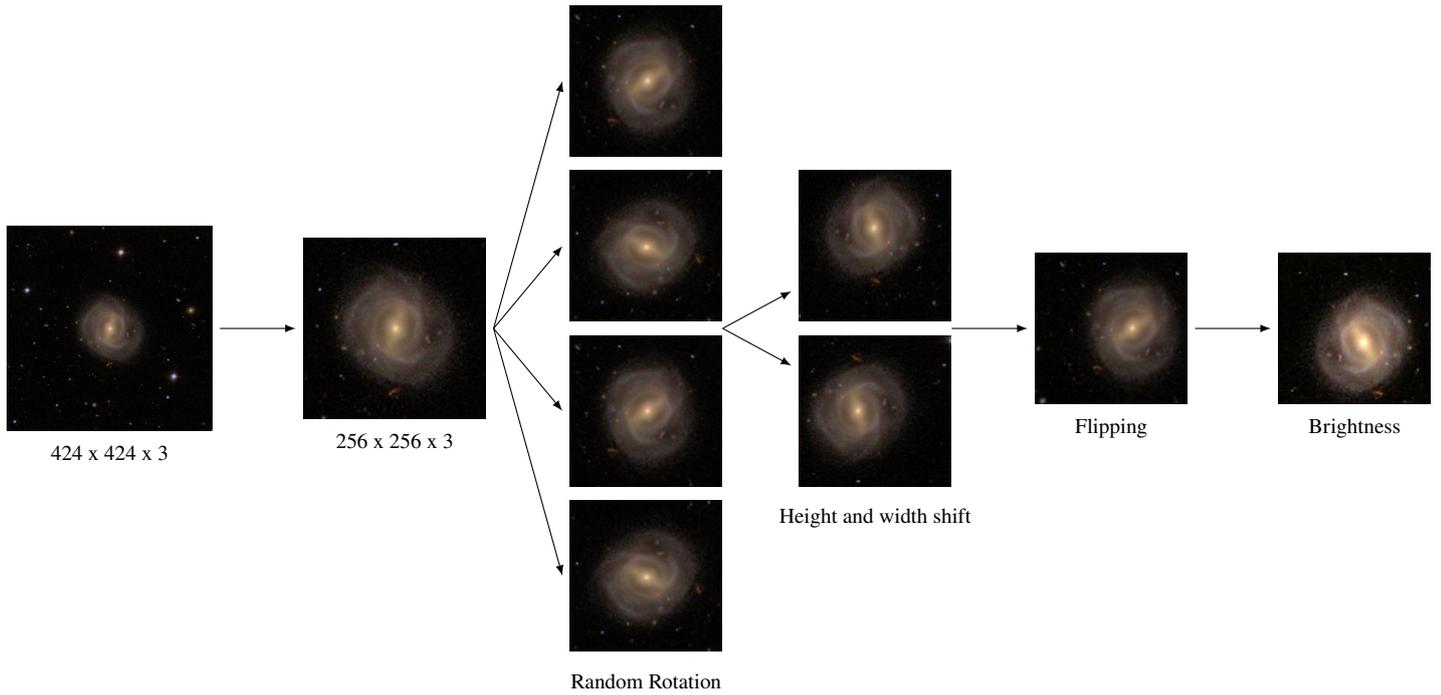}
    \caption{Image preprocessing. The original image is first centrally cropped and then randomly rotated in range [$0^\circ$, $100^\circ$]. After random rotation, it is randomly shifted vertically and horizontally. Then, vertical and horizontal flipping is applied before changing the brightness in a range of (0.9, 1.2) and passing the image into the network}
    \label{fig: Data_augment}
\end{figure*}
\subsection{Regression model}
\hspace{0.25 in}We implement a model that extracts the vote fraction probabilities of the 37 questions listed in the GZ2 decision tree (Table \ref{GZ2_DecisionTree}). The model will then be used to evaluate the output based on the test data and predict the probabilities of vote-fractions for the 37 questions in the GZ2 decision tree. \citet{Dieleman2015} proposed a model which performed with 7.466\% rmse score on the test set, which tops the public leader board of the competition conducted by Kaggle. The proposed model obtains results similar to it and obtains the third position with a public score of 7.765\%. 
\subsubsection{Data Preprocessing}
\hspace{0.25 in}As seen in the dataset, the images are actually composed of a large field view of the telescope with the galaxy as the center of interest. Hence, we need to crop the image. In practice, we centrally crop the image with a central window by dividing the image into 16 equal parts. The image is then cropped to a window of the central four sections and the remaining 12 sections are discarded as noise. This step is essential because it allows the main information to be contained in the centre of the image and eliminates all random noises like some other secondary object. 
\subsubsection{Data Augmentation}
\hspace{0.25 in}Data augmentation is a strategy that enables us to significantly increase the diversity of the data available. Data augmentation is an effective way of avoiding overfitting the network. Due to the limited data set, we use data augmentation to increase the size of training data. We use four different types of image augmentation. \\
\hspace{0.25 in}Firstly, we re-scale the image from [0, 255] to [0,1]. Then the image is randomly rotated in a range of 0$^\circ$ to 90$^\circ$. This is the first form of augmentation we apply. The second form of augmentation is translation. The randomly rotated image is translated horizontally or vertically by a factor of 10\%. This already increases the size of the training data. After the translations and rotations, the image is flipped. This form of augmentation alone allows the training data to be increased by a factor of 4. The horizontal and vertical flipping introduces even more diversity to the augmented data.\\
\hspace{0.25 in}Finally, the image brightness is adjusted in a scale of (0.9, 1.2) so as to optically distort the image. The final output is an image of size $224\times224\times3$ ($256\times256\times3$ for EfficientNetsB4-B7).
\subsection{Classification Model}
\hspace{0.25 in}The classification model has a similar preprocessing and augmentation to the regression model. The only difference is the input image size which is increased to $256\times256\times3$ because of the reduced dataset. We apply different efficientnet architectures and determine which work best for our particular problem. 
\citet{dai2018galaxy} proposed a model which performed with an F1 score of 0.9515. Our best performing model (EfficientNetB5) further improves it to 0.8857.
As mentioned in Table \ref{Classes}, we classified the dataset into 7 classes and the classification models were trained on 25,941 images in the dataset thus created. The following section will briefly discuss the 8 EfficientNets and the best performing architecture amongst them.
\subsection{Efficient Nets and compound scaling}
\hspace{0.25 in} \citet{efficientnet} proposed a simple mobile-size baseline architecture called the EfficientNet-B0 Architecture which made use of the squeeze and excite block and also employed a compound scaling method for increasing the model size to achieve better accuracy. The squeeze and excitation optimization was added to MBConv which are basically inverted residual blocks.\\
\hspace*{0.25 in}Increasing the depth of the neural network suffers from the famous vanishing gradient problem, even techniques like skip connections only offer slight benefit and an eventual saturation of the accuracy is observed. Similarly, increasing width or higher image resolution results in marginal accuracy gains. This points to the fact that any attempt to deal with this problem should change all these dimensions as a combination.\\
The Convolutional Neural Network can be expressed by the expression:
\begin{equation}   
    N(d,w,r)= \underset{1...S}{\odot}  \hat{F}_{i}^{d.\hat{L}_{i}} (X_{<r.\hat{H}_{i} , r.\hat{W}_{i} , \omega.\hat{C}_{i} >} )
\end{equation}
where N depicts the network, i represents the stage number. d , w , r are coefficients for scaling network width, depth and resolution; ${ \hat{F}_{i} }$ , ${ \hat{L}_{i} }$ , ${ \hat{H}_{i} }$ , ${ \hat{W}_{i} }$ , ${ \hat{C}_{i} }$  are predefined parameters in the baseline network.\\

The compound scaling technique uses the compound coefficient $\phi$ to uniformly scale network width, depth, and resolution in a principled way:

\begin{equation}
    \begin{array}{l}
        depth: d=\alpha^{\phi} \\ 
        width: w=\beta^{\phi} \\ 
        resolution : r=\gamma^{\phi} \\ 
      \hspace{0.25 in}  s.t \hspace{1cm} \alpha.\beta^{2}.\gamma^{2} \approx 2  \\
      \hspace{0.25 in} \hspace{1.3cm} \alpha\geq1, \beta\geq1, \gamma\geq1 
    \end{array}
\end{equation}
$\phi$ is a user-defined, global scaling factor that facilitates how many resources are available for model scaling whereas $\alpha$, $\beta$, and $\gamma$ determine how to assign these resources to network depth, width, and resolution respectively. The FLOPS (Floating Point Operations Per Second) of a convolutional operation are proportional to  $d$, $w^2$, $r^2$. So ,scaling the network using equation 3 will increase the total FLOPS by ($\alpha \times \beta^{2} \times \gamma^{2}) ^ \phi$ . Therefore, in order to make sure that the FLOPS don’t exceed the value of $2^\phi$, the constraint of $\alpha.\beta^{2}.\gamma^{2} \approx$ 2  is applied. Using Grid search by setting $\phi$ =1, the paramters $\alpha$, $\beta$ , $\gamma$ can be determined that result in the best accuracy. Once $\alpha$, $\beta$ , $\gamma$ are determined, compound coefficient $\phi$ can be increased to get larger but more accurate models. Thus keeping the baseline architecture EfficientNet-B0, EfficientNet-B1 to EfficientNet-B7 were created where the last number represents the value of $\phi$.

\subsection{Network architecture}
\begin{figure}
    \centering
    \includegraphics[width=0.3\textwidth, height=0.5\textwidth]{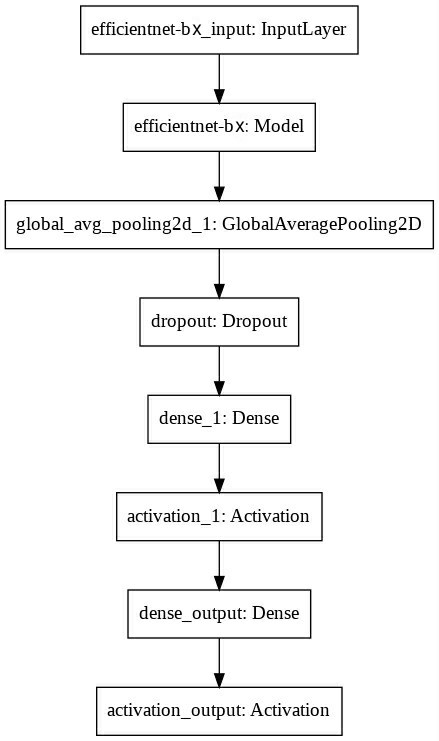}
    \caption{Network architecture of the tail part of the fine-tuned model. The \textit{x} in \textit{efficientnet-bx} is a placeholder for all the EfficientNet architectures ranging between B0-B7}
    \label{fig:Network Architecture}
\end{figure}

\hspace*{0.25 in} We initially tried using sequential convolutional networks as an approach to the regression problem. Sequential convolutions have a drawback of limited depth. As the main motive was feature extraction, it was better to go with ResNets \citep{he2016}. The Residual blocks, as shown in section 3.3, fig. \ref{fig: residual block}, seemed to solve most of the problems and boosted the accuracy by two-folds. But it still wasn't enough as other network architectures such as \citet{Dieleman2015} with 7.74\% rmse for the regression problem and \citet{dai2018galaxy} with about 95\% accuracy for the classification problems performed substantially better. To overcome this resistance, a new architectural block, the Squeeze and Excitation block as shown in fig.\ref{SE Block}, was implemented with custom network architectures. This increased the performance of both the classification and the regression model enabling us to achieve an rmse score of 9.110\% on the submission on kaggle. The customized models were built to keep the network concise and short but were too shallow. \citet{he2016} proved that deeper networks perform well with feature extractions. We decided to use transfer learning and a ready made architecture called the EfficientNets. Pretrained weights from imagenet were chosen to be used because the network was benefited by weights pre-trained on the vast imagenet dataset with varying shapes that proved useful in the detection of basic galaxy shapes. For examples, the detection of ellipses of elliptial galaxies or spirals for spiral galaxies. \citep{efficientnet}.\\
\hspace*{0.25 in}The EfficientNet architecture uses compound scaling, refer section 4.3, to improve performance. We use the EfficientNet architectures for both the models and introduce a tail part that helps fine-tune the model. The tail part of the model takes the output of the EfficientNet and introduces a Global Average Pooling which downsamples the output of every dimension and flattens the tensor. A dropout is introduced with rate of 0.5 to reduce overfitting and prevent complex co-adaptations on training data. The output of this non-linear transformation is then passed onto a FC layer with 64 connections with a \textit{ReLU} activation to increase non-linearity. The output layer consists of 7 classes which use a \textit{softmax} activation. The regression model uses the same tail part with slight modifications to the dropout rates and output layer. We use Adam optimizer with dynamically varying learning rate which drops when the loss plateaus. Figure \ref{fig:Network Architecture} shows a brief overview of how the network architecture is modelled for the tail part.\\

\subsection{Implementation Details}
\hspace*{0.25in}We use a mini-batch gradient descent with a varying batch size for each of the EfficientNets. We begin with a batch size of 256 and reduce it down to 64 till the last network. We use Adam optimizer with $\beta_1 = 0.9 , \beta_2 = 0.999$. The initial learning rate is set to 1.5e-4 and it reduces by a factor of 0.2 on a plateau with patience of 4. We train our models over 50 epochs saving the best model on every epoch end based on the validation loss. We use early stopping with a patience of 9. The initial image size is $224\times224$ for B0-B3 but the size is increased to $256\times256$ in the later EfficientNets. The dropout probability is set to 0.5. The weights are initialized using He initialization. Our implementation is using Python, pandas, numpy, scikit-learn, Tensorflow and Keras. It takes about 10 to 35 hours to train EfficientNets B0-B7 over 50 epochs on a NVIDIA Tesla K80 GPU.

\section{Results and Discussions}
\hspace*{0.25in}In this section we describe the performance metrics used to evaluate the performance of both the models. The metrics used are rmse, accuracy, precision, recall, F1 score and confusion matrix. The results of the 8 architectures are compared based on their accuracy, precision, recall and F1 classification report for the classification model and the rmse score for the regression model. We present these results to determine which network architecture works best for our experimental setup and provide the confusion matrix of the best performing network on the classification model. 

\subsection{Performance Metrics}
\subsubsection{Accuracy}
\hspace*{0.25 in} Accuracy can be defined as the fraction of total samples which our model correctly predicted against the total number of samples. It can be formally defined as:
\begin{equation}
\text{Accuracy} = \frac{\text{Number of correct Predictions}}{\text{Total number of Predictions}}
\end{equation}

\subsubsection{Precision}
\hspace*{0.25 in} Precision is used to measure the performance when the cost of false positives is high. It answers the question 'What proportion of positive identifications were actually correct?' It can be formally defined as: 
\begin{equation}
\text{Precision} = \frac{\text{True Positives}}{\text{True Positives + False Positives}}
\end{equation}

\subsubsection{Recall}
\hspace*{0.25 in} Recall is used to measure the performance when the cost of false negatives is high.  It answers the question 'What proportion of actual positives were correctly predicted?' It can be formally defined as: 
\begin{equation}
\text{Recall} = \frac{\text{True Positives}}{\text{True Positives + False Negatives}}
\end{equation}

\subsubsection{F1 Score}
\hspace*{0.25 in} F1 Score is the overall measure of of a model's accuracy, it is the harmonic mean of precision and recall. A higher F1 score indicates that you have both low false positives and false negatives. It can be formally defined as: 
\begin{equation}
\text{Recall} = 2\times \frac{\text{Precision} \times \text{Recall}}{\text{Precision + Recall}}
\end{equation}

\subsubsection{Confusion Matrix}
\hspace*{0.25 in} It is the table which shows the Predicted vs Actual Classifications. Formally, it can be defined as :
\begin{equation}
CM_{xy}(x,y=1,2...n_{samples})
\end{equation}
where each entry is the number of true classes x, but predicted to y.

\subsubsection{RMSE Score}
\hspace*{0.25 in} Root mean squared error (RMSE) is the square root of the mean of the square of all of the error. It is the sample standard deviation of the differences between predicted values and actual values.
\begin{equation}
\sqrt{\frac{ 1 }{ n } \sum_{ i = 1 } ^ { n } ( Predicted_{i}-Actual_{i} ) ^ 2 }
\end{equation}

\subsection{Regression Results and Discussion}
\hspace*{0.25 in}In this section we present the results of the Regression model by comparing the performance of all eight EfficientNets over \textit{rmse} metric. Table \ref{tab: regression_results} shows a summary of the rmse score obtained for regression models over test set for all EfficientNets.
The model outputs have been evaluated by submitting to the Kaggle Galaxy Zoo challenge and the scores for output files have been documented. We can observe that an ensemble of models B0 and B2 has given promising results and we were placed \textbf{third} on the public leaderboard.
\begin{center}
    \begin{table}
    \centering
    \begin{tabular}{c | c c}
        \hline
        \hline
        \multirow{1}{*}{EfficientNets} & 
            \multicolumn{1}{c}{RMSE Score} \\
        \hline
        EfficientNetB0 & 0.08099  \\
        EfficientNetB1 & 0.09759  \\
        EfficientNetB2 & \textbf{0.08086}  \\
        EfficientNetB3 & 0.08447  \\
        EfficientNetB4 & 0.08174  \\
        EfficientNetB5 & 0.08575  \\
        EfficientNetB6 & 0.08793  \\
        EfficientNetB7 & 0.08884  \\
        Averaging over EfficientNet B0 and B2 & \textbf{0.07792} \\
        \hline
    \end{tabular}
    \caption{Results of Regression model using EfficientNets from B0-B7}
    \label{tab: regression_results}
\end{table}
\end{center}
\subsection{Classification Results and Discussion}
\hspace*{0.25 in}This section summarizes the results obtained by the classification model on different performance metrics namely accuracy, precision, recall and F1 score. We compare the performance of the best performing EfficientNet with other state of the art networks such as the VGG19, ResNet50 and InceptionV3 along with the network proposed by \citet{dai2018galaxy}.\\
\begin{table}
    \centering
    \begin{tabular}{c | c c c c}
        \hline
        \hline
        \multirow{2}{*}{EfficientNets} & 
            \multicolumn{4}{c}{Performance}\\& Accuracy
            & Precision & Recall & F1 Score \\
        \hline
        EfficientNetB0 & 0.9219 & 0.8714 & 0.8742 & 0.8714 \\
        EfficientNetB1 & 0.9293 & 0.8842 & 0.8842 & 0.8828 \\
        EfficientNetB2 & 0.9331 & 0.8828 & 0.8814 & 0.8814 \\
        EfficientNetB3 & 0.9347 & 0.8871 & 0.8814 & 0.8814 \\
        EfficientNetB4 & 0.9324 & 0.8842 & 0.8785 & 0.8828 \\
        EfficientNetB5 & \textbf{0.9370} & 0.8885 & \textbf{0.8900} & \textbf{0.8857} \\
        EfficientNetB6  & 0.9335 & \textbf{0.8900} & 0.8614 & 0.8742\\
        EfficientNetB7 & 0.9208 & 0.8771 & 0.8642 & 0.8671 \\
        \hline
    \end{tabular}
    \caption{Classification Results for EfficientNet B0-B7}
    \label{tab:classification_results}
\end{table}
\hspace*{0.25 in}Table \ref{tab:classification_results} summarizes the performance metrics of all the efficientnets over the test set. The 8 networks achieve an accuracy of 0.9219, 0.9293, 0.9331, 0.9347, 0.9324, 0.9370, 0.9335 and 0.9208 respectively. We observe that with our setup, EfficientNetB5 performs the best on the classification problem.\\
\hspace*{0.25 in}The class-wise metrics of EfficientNetB5 are shown in \ref{tab:B5 Class report}. It can be observed that the network has a slight problem adjusting to the samples in Cigar shaped smooth (class 2). The limited amount of training data may have caused the network to not generalize well on class 2. 

\begin{center}
    \begin{table}
    \centering
    \begin{tabular}{c | c c c}
        \hline
        \multirow{2}{*}{Class} & 
            \multicolumn{3}{c}{Performance}\\
            & Precision & Recall & F1 Score\\
        \hline
        0 & 0.96   &   0.97  &    0.97\\
        1 & 0.94   &   0.97  &    0.96\\
        2 & 0.62   &   0.72  &    0.67\\
        3 & 0.97   &   0.91  &    0.94\\
        4 & 0.95   &   0.85  &    0.90\\
        5 & 0.91   &   0.93  &    0.92\\
        6 & 0.87   &   0.80  &    0.84\\
        Average & \textbf{0.8885} & \textbf{0.8900} & \textbf{0.8857}\\
        \hline
    \end{tabular}
    \caption{Class Performance for EfficientNetB5}
    \label{tab:B5 Class report}
\end{table}
\end{center}

Amongst all the classes, Lenticulars (class 3) performed best in the precision for EfficientNetB5. Even though the precisions of all the EfficientNets were comparably similar, it can be observed that EfficientNetB6 has a slightly higher precision. But based on the F1 score, we can conclude that EfficientNetB5 has a greater performance overall. According to the classification report of EfficientNetB5 as shown in Table \ref{tab:B5 Class report}, Lenticulars (class 3) have shown to have the best precision of 0.97. Completely Round Smooth (class 0) has the next best precision, 0.96, followed by Barred Spirals (class 4) with precision 0.95.\\ 
\hspace*{0.25 in}The Recall for all the 8 networks shows the same behaviour as precision. EfficientNetB5 again shows the best performance in regards to the recall. It can be observed that Completely Round Smooth (class 0) and In-Between Smooth (class 1) have almost identical Recall for B5 of 0.97. It's followed by a recall value of 0.93 which is for Unbarred Spirals (class 5). The results show consistency only in one class i.e. Completely Round Smooth galaxies (class 0). We can speculate the reason for this to be the sheer volume of training data available for class 0. The class wise F1 score for B5 shows that the network performs best on the Completely Round Smooth samples (class 0) followed by In-between smooth samples (class 1). Again, we speculate the reason for this behaviour to be the large volume of data available which makes our training and testing data set quite skewed in nature.\\
\begin{table}
    \centering
    \begin{tabular}{c|c c c c c c c}
         & 0 & 1 & 2 & 3 & 4 & 5 & 6 \\
         \hline
         0 &\textbf{784} &22 &0 &0  &0 &1 & 3 \\
         1 &15 &\textbf{758} &2 &0  &0 &2 & 1 \\
         2 &0 &5 &\textbf{41} &11  &0 &0 & 0 \\
         3 &0 &3 &22 &\textbf{344}  &0 &2 & 7 \\
         4 &0 &0 &0 &1  &\textbf{70} &10 & 1 \\
         5 &8 &8 &1 &0  &1 &\textbf{304} & 6 \\
         6 &6 &8 &0 &0  &3 &14 & \textbf{125} \\
    \end{tabular}
    \caption{Confusion Matrix: EfficientNetB5}
    \label{tab:CF B5}
\end{table}
\hspace*{0.25 in}Table \ref{tab:CF B5} shows the confusion matrix of EfficientNetB5. It can be observed that the network misclassifies 16 of the samples out of the total 57 testing samples. These images are mostly classified into In-between smooth (class 1) and Lenticulars (class 3) which are quite similar to Cigar-shaped smooth samples (class 2). This misclassification occurs in all of the networks so we can assume that a larger training set may solve the problem in the future. The network also seems to have wrongly classified 12 samples of Barred Spirals (class 4) out of the 82 test images, resulting in lower recall. Again, the misclassified images are wrongly classified into the next following class, i.e. Unbarred Spirals (class 5) which are very similar to the samples from Barred Spirals (class 4) and are very difficult to even distinguish visually. Most of the misclassifications may have occurred because of the similarities between the classes, e.g. Cigar-shaped galaxies and Lenticulars. \\
\hspace*{0.25 in}We compare the performance of EfficientNetB5 with different state of the art networks. \citet{dai2018galaxy} proposed a network for classifying the galaxies into 5 different classes. We extend the architecture to classify the galaxies into 7 classes. Accordingly, we observe that the performance of the model reduces drastically by the addition of 2 classes. The network gave an accuracy of 95.20\% over 5 classes but the accuracy reduces down to 57.12\% when we add two additional nodes into the output layer. 
\begin{table}
    \centering
    \begin{tabular}{c | c c c c}
        \hline
        \hline
        \multirow{2}{*}{Networks} & 
            \multicolumn{4}{c}{Performance}\\
            & Accuracy & Precision & Recall & F1 Score\\
        \hline
        VGG-16 & 0.7547 &  0.6043 & 0.5714 & 0.5771\\
        Resnet-50 & 0.8323 & 0.7800 & 0.7114 & 0.7328\\
        J.M. Dai Model & 0.5712 & 0.4271 & 0.4214 & 0.3828\\
        InceptionV3 & 0.9347 & 0.8785 & 0.8871 & 0.8771 \\
        \textbf{EfficientNetB5} & 0.9370 & 0.8885 & 0.8900 & 0.8857\\
        \hline
    \end{tabular}
    \caption{Performance comparision over different networks}
    \label{tab:Comparision}
\end{table}

The ResNets proposed by \citet{he2016} showed better results than VGG-16 by giving an accuracy of 83.23\%. The same identity mappings and residual connections have been used in EfficientNets so it's not surprising that they perform better \citep{tan2019efficientnet}. The VGG-16 shows decent results over the 7 class classification but still performs poorly because of a very low F1 score. We suppose the cause for the poor performance is due to the skewness of the dataset. The InceptionV3 shows results comparable to the Efficient Nets with an accuracy of 93.47\% and an F1 score of 0.8771.

\subsection{Feature Map Visualization}
\hspace*{0.25in} The computational workings of a CNN are opaque to humans and are considered to be difficult to grasp, therefore CNNs are classified as 'black boxes'. We wanted to implement a visualization technique that gives insight into the function of intermediate feature layers and the operation of the classifier network \citep{zeilar2014}. We extract the 4 intermediate layers from a single block of the Resnet-50 Architecture.

\begin{figure}
    \centering
    \includegraphics[width=0.3\textwidth, height=0.5\textwidth]{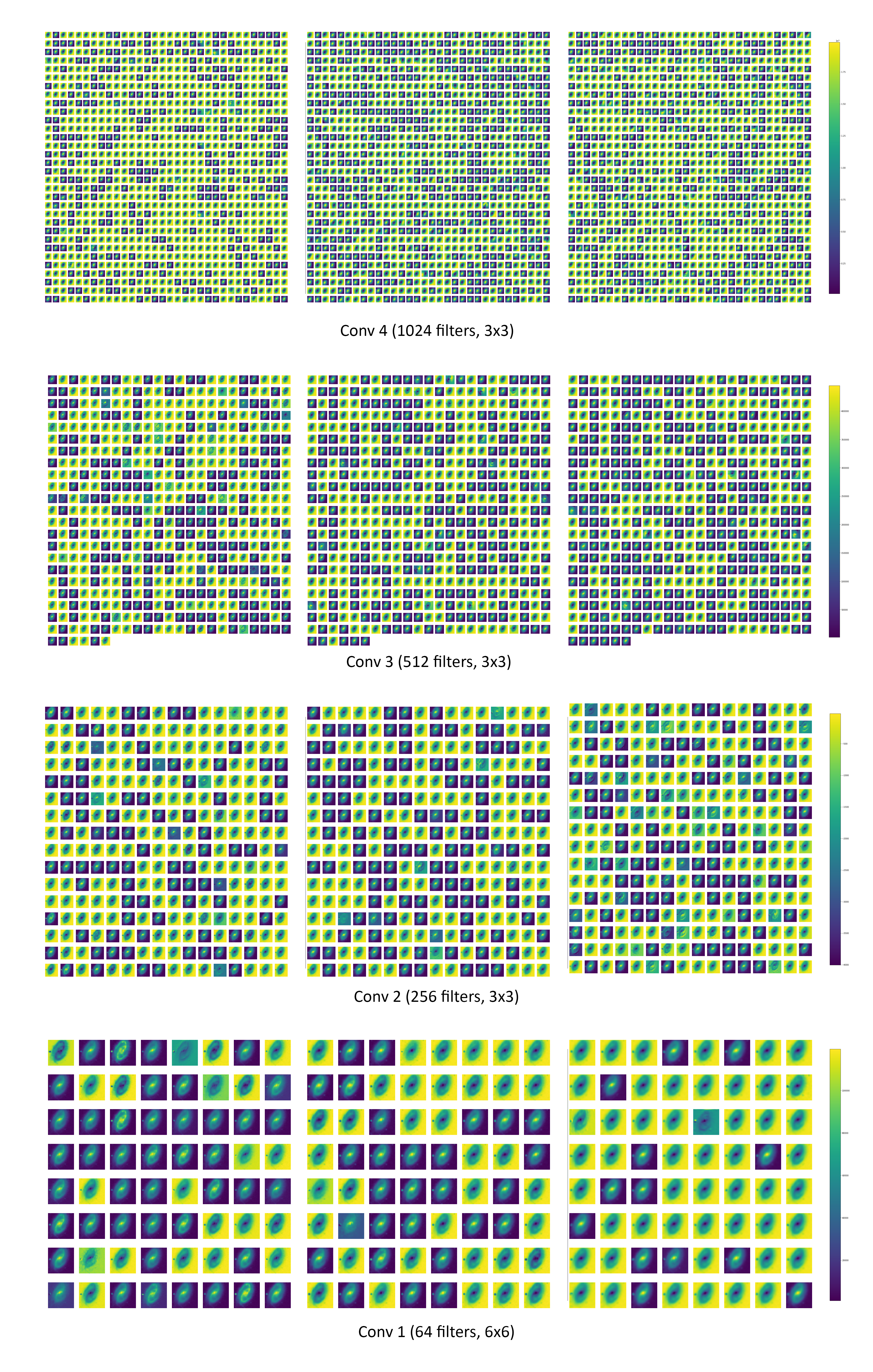}
    \caption{Filter Visualization of convolutional layers, from Top to Bottom ie Conv4, Conv3, Conv2, Conv1. Each filter is of size $3\times3$ and randomly sampled activations from left to right are represented here. The third image represents activations after residual mapping operations}
    \label{fig:Filter Visualization}
\end{figure}

\hspace*{0.25 in} In Figure \ref{fig:Filter Visualization} we can observe in the first layer, that galaxy edges and corners are being detected. In the 64 filters used, we can see that different channels ie RGB are partial to specific filters. As a result, we can see certain filters enhanced each of the colour channel. The first layer of outputs, do not extract any complex or visually distinguishable features. In the second layer, with an increased number of filters, we can see a decrease in emphasis on the colour of the galaxy owing to elimination of the RGB colour channels. Due to an increase in the number of channels, we can observe specialization in feature extraction as opposed to the corners and edges, one may observe that expanding the filters results in synergizing  the data learned from previous layers and activations of subtle shapes such as ellipse, circle, etc. The third layer is where we can see abstract shapes which are more prominent than the previous activations. Details of the galaxy such as outline of the arms in case of spirals and a bulge in case of ellipticals can be seen. In the last layer, we can observe a vivid difference between features extracted, which is what we expect from a classification model. Complex shapes such as the number of spiral arms or barred/unbarred cores can be approximated from such feature representations. \\
\hspace*{0.25 in}It is evident from the visualized feature maps that progressive expansion in the number of filters results in a growing complexity of the representations in the underlying features of the image. As observed in the third image of every row, the residual connections ensure elimination of weak representations from the layer outputs. 

\section{Conclusion}
\hspace*{0.25 in}In this paper, we explore the usage of EfficientNets for Galaxy Morphology Classification. We classify 25,941 galaxies in 7 classes, namely Completely Round Smooth, In-Between Smooth, Cigar-shaped smooth, Lenticulars, Barred Spirals, Unbarred Spirals and Irregular using the Galaxy Zoo 2 dataset.  We attempt at predicting the vote fractions of the 79,975 testing images from the same data release in the data from the original Galaxy Zoo 2 challenge on Kaggle. We perform data preprocessing using a complete preprocessing pipeline with 4 different augmentation techniques to avoid overfitting. The EfficientNets use the residual mappings from \citet{he2016} and the Squeeze \& Excitation blocks from \citet{hu2018squeeze}. Coupled with compound scaling, it performs better than most of the networks created before. Among the EfficientNets, we observe that EfficientNetB5 performs better than other EfficientNets for classification and EfficientNetB0 and EfficientNetB2 performs the best for regression (vote fraction predictions).

For future work, we would like to experiments with combination of Efficient Nets with other state of the art architectures. We would like to train these networks on even larger datasets.

In future, many large scale surveys such as the Dark Energy Survey (DES), Large Synoptic Space Telescope (LSST), Australian Square Kilometre Array Pathfinder (ASKAP), will obtain billions of images of galaxies and usage of EfficientNets may be applied to automatically classify the galaxies to achieve better and faster performance. 

\section{Acknowledgements}
\hspace*{0.25 in}We thank the Galaxy Zoo challenge, SDSS and Kaggle platform for providing the data. We thank Google for letting us use their Colaboratory platform. We would also like to thank Mr. Revanth Regeti for valuable contribution to the code for our experimental setup.

\section{Data Availability}
Our code is available at \href{https://github.com/obi-wan-shinobi/GalaxyEfficientNets}{https://github.com/obi-wan-shinobi/GalaxyEfficientNets}. The dataset can be downloaded from the official Kaggle website under the Galaxy Zoo 2 competition. For the regression model, the data of 61,578 images is used without any alterations. For the classification problem, the data has 25,941 images and is derived from the competition training data and can be obtained by following the vote fractions thresholds mentioned in Table \ref{Dataset}. The dataset can also be downloaded from the github repository mentioned above. 



\bibliographystyle{mnras}
\bibliography{Main_Paper} 





\bsp	
\label{lastpage}
\end{document}